\documentclass[10pt,journal,compsoc]{IEEEtran}
\ifCLASSOPTIONcompsoc
\usepackage[nocompress]{cite}
\else
\usepackage{cite}
\fi
\ifCLASSINFOpdf
\else
\fi

\usepackage{framed,multirow}

\usepackage{amssymb}
\usepackage{latexsym}
\usepackage{amsmath}
\usepackage{url}
\usepackage{xcolor}
\usepackage{algorithmic}
\usepackage{graphicx}
\usepackage{textcomp}

\usepackage{algorithm}

\usepackage{subfigure}
\usepackage{booktabs}  

\definecolor{newcolor}{rgb}{.8,.349,.1}

\begin{document}
	%
	\title{Deep Multi-task Multi-label CNN for Effective Facial Attribute Classification}
	%
	%
	%
	%
	
	\author{Longbiao~Mao, Yan~Yan,~\IEEEmembership{Member,~IEEE}, Jing-Hao Xue, and Hanzi~Wang,~\IEEEmembership{Senior Member,~IEEE}
		\IEEEcompsocitemizethanks{\IEEEcompsocthanksitem Corresponding author: Yan Yan.\protect
			
			\IEEEcompsocthanksitem L.~Mao, Y.~Yan, H.~Wang are with the Fujian Key Laboratory of Sensing and Computing for Smart City, School of Informatics, Xiamen University, Xiamen 361005, China
			(email: maolongbiaocool@qq.com; yanyan@xmu.edu.cn; hanzi.wang@xmu.edu.cn).\protect
			\IEEEcompsocthanksitem J.-H. Xue is with the Department of Statistical Science, University College London, London WC1E 6BT, UK (e-mail: jinghao.xue@ucl.ac.uk).\protect
		}
		}

	%
	%

	\markboth{Journal of \LaTeX\ Class Files}
	{Shell \MakeLowercase{\textit{et al.}}: Deep Multi-task Multi-label CNN for Effective Facial Attribute Classification}
	%



	\IEEEtitleabstractindextext{%
		\begin{abstract}
			Facial Attribute Classification (FAC) has attracted increasing attention in computer vision and pattern recognition. However, state-of-the-art FAC methods perform face detection/alignment and FAC independently. The inherent dependencies between these tasks are not fully exploited. In addition, most methods predict all facial attributes using the same CNN network architecture, which ignores the different learning complexities of facial attributes.
			To address the above problems, we propose a novel deep multi-task multi-label CNN, termed DMM-CNN, for effective FAC. Specifically,
			DMM-CNN jointly optimizes two closely-related tasks (i.e., facial landmark detection and FAC) to improve the performance of FAC by taking advantage of multi-task learning.
			To deal with the diverse learning complexities of facial attributes, we divide the attributes into two groups: objective attributes and subjective attributes. Two different
			network architectures are respectively designed to extract features for two groups of attributes, and a novel dynamic weighting scheme is proposed to automatically assign the loss weight to each facial attribute during training.
			Furthermore,  an adaptive thresholding strategy is developed to effectively alleviate the problem of class imbalance for multi-label learning.
			Experimental results on the challenging CelebA and LFWA datasets show the superiority of the proposed DMM-CNN method compared with several state-of-the-art FAC methods.
		\end{abstract}
		\begin{IEEEkeywords}
			facial attribute classification, multi-task learning, multi-label learning, convolutional neural network
		\end{IEEEkeywords}}
		\maketitle

		\IEEEdisplaynontitleabstractindextext

		%
		\IEEEpeerreviewmaketitle

\section{Introduction}\label{sec:introduction}

\begin{figure}[!t]
\centerline{\includegraphics[width=8cm, height=6.5cm]{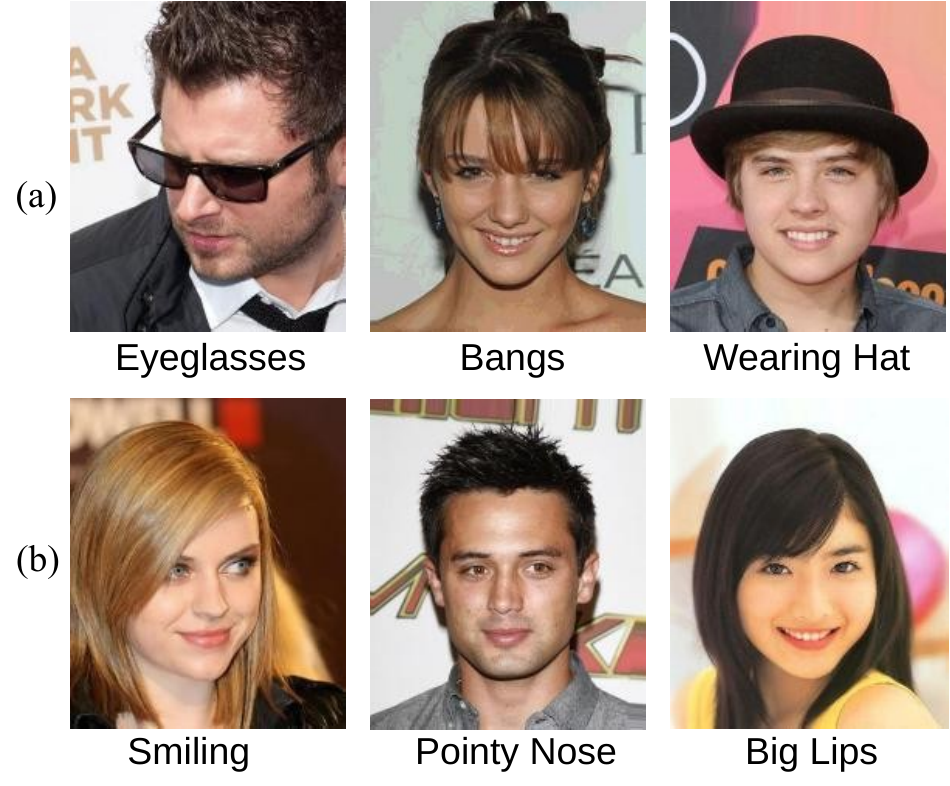}}
\caption{Examples of different facial attributes. (a) Objective attributes: Eyeglasses, Bangs and Wearing Hat; (b) Subjective attributes: Smiling, Pointy Nose and Big Lips.}
\label{fig:sample}
\end{figure}

%
%
%
%
During the past few years, Facial Attribute Classification (FAC) has attracted significant attention in computer vision and pattern recognition, due to its widespread applications, including image retrieval \cite{b1, b2}, face recognition \cite{b3, b4}, person re-identification \cite{b5, b6}, micro-expression recognition \cite{b7}, image generation \cite{b8} and recommendation systems \cite{b9, b10}. Given a facial image, the task of FAC is to predict multiple facial attributes, such as gender, attraction and smiling (some facial attributes are shown in Fig.~\ref{fig:sample}).
Although the task of FAC is only an image-level classification task, it is not trivial, mainly because of the variability of facial appearances caused by significant changes in viewpoint, illumination, etc.


Recently, due to the outstanding performance of Convolutional
Neural Network (CNN), most state-of-the-art FAC methods take advantage of CNN to classify facial attributes. Roughly speaking, these methods can be categorized as follows: (1) single-label learning based FAC methods  \cite{b11, b12, b13} and (2) multi-label learning based FAC methods \cite{b14, b15, b16, b17, b18}. The single-label learning based FAC methods  usually extract the CNN features of facial images and then classify facial attributes by the Support Vector Machine (SVM) classifier. These methods, however, predict each attribute individually, thus ignoring the correlations between attributes. In contrast, multi-label learning based FAC methods, which can predict multiple attributes simultaneously, extract the shared features from the lower layers of CNN and learn attribute-specific classifiers on the upper layers of CNN.

Typically, the
above methods firstly perform face detection/alignment and then predict facial attributes. In other words, these closely-related tasks are trained separately. Therefore, the intrinsic relationships between these tasks are not fully and effectively exploited. Moreover, some multi-label learning based FAC methods (such as \cite{b19, b20}) are developed to simultaneously predict facial attributes by using a single CNN. These methods treat the diverse attributes equally (using the same network architecture for all attributes), ignoring the different learning complexities of these attributes (for example,
learning to predict the ``Wearing-Eyeglasses" attribute may be much easier than identifying the ``Pointy Nose" attribute, as shown in Fig.~1).
In particular, some attributes (e.g., ``Big Lips", ``Oval Face") are very subjective, and they are more difficult to be recognized and may even confuse humans sometimes.
Even worse, the training set often suffers from the problem of imbalanced labels for some facial attributes (e.g., the ``Bald" attribute has very few positive samples). Re-balancing multi-label data is not a trivial task.



To alleviate the above problems, we propose a novel Deep Multi-task Multi-label CNN method (DMM-CNN) for effective FAC. Two closely-related tasks (i.e., Facial Landmark Detection (FLD) and FAC) are jointly optimized to boost the performance of FAC based on multi-task learning. As a result, by exploiting the intrinsic relationship between the two tasks, the performance of FAC is effectively improved. Considering the diverse learning complexities of facial attributes, we divide the facial attributes into two groups: objective attributes and subjective attributes, and further employ two different network architectures to respectively extract discriminative features for these two groups.
We also develop a novel dynamic weighting
scheme to dynamically assign the loss weights to all facial attributes during training.
Furthermore, in order to alleviate the problem of class imbalance for multi-label training, we develop an adaptive thresholding strategy to effectively predict facial attributes.

Similar to our previous MCFA method \cite{b18}, the proposed DMM-CNN method also adopts the
framework of multi-task learning. However, there are several  significant differences between MCFA and DMM-CNN. Firstly, MCFA focuses on solving the problem of extracting semantic attribute information by using a multi-scale CNN, while DMM-CNN aims to overcome the problem of diverse learning complexities of facial attributes (by designing different network architectures for objective and subjective attributes, and proposing a dynamic weighting scheme). Secondly, MCFA uses a fixed decision threshold for all attributes, while DMM-CNN leverages an adaptive thresholding strategy to alleviate the problem of class imbalance. Thirdly,  MCFA jointly learns the tasks of face detection, facial landmark detection (FLD) and FAC,
while DMM-CNN simulteneously performs FLD and FAC. The reason why face detection is not adopted in DMM-CNN is that using the auxiliary task of face detection only slightly improves the performance of FAC, but significantly increases the computational burden. Moreover, FLD explicitly plays the role of face localization.
Finally, the FLD module in MCFA only gives five off-the-shelf facial landmarks (left and right eyes, the mouth corners, and the nose tip). In contrast, the FLD module in DMM-CNN outputs 72 facial landmarks, which can provide more auxiliary information beneficial for FAC.

The main contributions of this paper are summarized as follows:
\begin{itemize}

\item We divide the diverse facial attributes into objective attributes and  subjective attributes according to their different learning complexities, where two different levels of SPP layers (i.e., a 1-level SPP layer and a 3-level SPP layer) are used to extract features.  To the best of our knowledge, this paper is the first work to learn multiple deep neural networks to enhance the performance of FAC by considering the different learning complexities of facial attributes (objective and subjective attributes).

\item A novel dynamic weighting scheme, which capitalizes on the rate of validation loss change
obtained from the whole validation set, is proposed to automatically assign  weights to facial attributes.
     In this way, the training process concentrates on classifying the more difficult facial attributes.

\item We develop  an adaptive  thresholding strategy to accurately classify facial attributes for multi-label learning. Such a strategy takes into account the imbalanced data distribution of facial attributes. Thus, the problem of class imbalance for some attributes of FAC is effectively alleviated from the perspective of decision level.

\end{itemize}

The remainder of this paper is organized as follows. In Section 2, we review related work. In Section 3, we introduce the details of the proposed method. In Section 4, we evaluate the performance of the proposed method and compare it with several state-of-the-art methods on the challenging CelebA and LFWA datasets. Finally, the conclusion is drawn in Section 5.

\section{Related Work}

Over the past few decades, great progress has been made on FAC. Traditional FAC methods \cite{b3, b21} rely on hand-crafted features to perform attribute classification.
With the development of deep learning, current state-of-the-art FAC methods employ CNN models to predict the attributes and have shown remarkable improvements in performance. 
Our proposed method is closely related to CNN-based multi-task learning, multi-label learning and attribute grouping. In this section, we briefly introduce related work based on CNN.
\subsection{Multi-task Learning}
Multi-task Learning (MTL) \cite{b22} is an effective learning paradigm to improve the performance of a target task with the help of some related auxiliary tasks. MTL has proven to be effective in various computer vision tasks \cite{b23, b24, b25}. The CNN model can be naturally used for MTL, where all the tasks share and learn common feature representations in the deep layers.
 For example, Zhang et al. \cite{b26} perform FLD together with several related tasks, such as gender classification and pose estimation.
Tan et al. \cite{tan} jointly learn multiple attention mechanisms (including parsing attention, label attention and spatial attention) in an MTL manner for pedestrian attribute analysis.

Appropriately assigning weights to different loss functions plays an importance role for multi-task deep learning.
Kendall et al. \cite{kendall} propose to weigh loss functions based on the homoscedastic uncertainty of each task, where the weights are automatically learned from the data. Chen et al. \cite{chen} develop
a gradient normalization (GradNorm) method which performs multi-task deep learning by  dynamically tuning gradient magnitudes. The loss weights are assigned according to the training rates of different tasks. Recently, Liu et al. \cite{Liu} develop a multi-task attention network, which automatically learns both task-shared and task-specific features in an end-to-end manner, for MTL. They develop a novel weighting scheme, Dynamic Weight Average (DWA),
which learns the weights based on the rate of loss changes for each task.
%

\subsection{Multi-label Learning}
On one hand, traditional CNN based FAC methods mainly rely on single-label learning to predict facial attributes. For example, Liu et al. \cite{b27} propose to cascade two Localization Networks (LNets) and an Attribute Network (ANet) to localize face regions and extract features, respectively. They use the features extracted from ANet to train 40 SVMs to classify 40 attributes. The single-label learning based FAC methods consider the classification of each attribute as a single and independent problem, thereby ignoring the correlations among attributes. Moreover, these methods are usually time consuming and cost prohibitive.

On the other hand, multi-label learning based FAC methods predict multiple
facial attributes simultaneously in an end-to-end trained network. Because each face image is naturally associated with multiple attribute labels, multi-label learning is well suited for FAC.
For example, Ehrlich et al. \cite{b28} use a Restricted Boltzmann Machine (RBM) based model for attribute classification. Rudd et al. \cite{b19} introduce a Mixed Objective Optimization Network (MOON) to address the multi-label imbalance problem.
Huang et al. \cite{Huang} propose a greedy neural architecture search method to automatically discover the optimal tree-like network architecture, which can jointly predict multiple attributes.

Existing multi-label learning based FAC methods, which use the same network architecture for each attribute, usually learn the features of facial attributes on the upper layers of CNN.
However, different facial attributes have different learning complexities. Therefore, it is more attractive to develop a new CNN model, which considers the diverse learning complexities of attributes rather than treating the attributes equally during the training stage.

%
%

\subsection{Attribute Grouping}
Facial attributes can be divided into several groups according to different criteria.
For example, Emily et al. \cite{b20} divide the facial attributes into 9 groups according to the attribute location, and explicitly learn the relationships among attributes from similar locations in a face image. Han et al. \cite{b29} group the face attributes into ordinal and nominal attributes, holistic and local attributes in terms of data type and semantic meaning. Accordingly, four types of sub-networks (having the same network architecture) corresponding to the holistic-nominal, holistic-ordinal, local-nominal and local-ordinal attributes are defined, where a different loss function for each sub-network is used for FAC.
Cao et al. \cite{Cao} split the facial attributes into four attribute groups including upper, middle, lower, and whole image according to the corresponding locations and design four task specific sub-networks (corresponding to four attribute groups) and one shared sub-network for FAC.


 In this paper, different from the above attribute grouping methods, we propose to divide the attributes into two groups: objective attributes and subjective attributes based on their different learning complexities. Accordingly, we design two different network architectures, which are able to extract different levels of features beneficial to classify objective and subjective attributes, respectively.

%

\section{Methodology}

In this section, we introduce in detail the proposed DMM-CNN method, which takes advantage of multi-task learning and multi-label learning, for effective FAC.

\subsection{Overview}

\begin{figure*}[!t]
\centering
\includegraphics[width=5in]{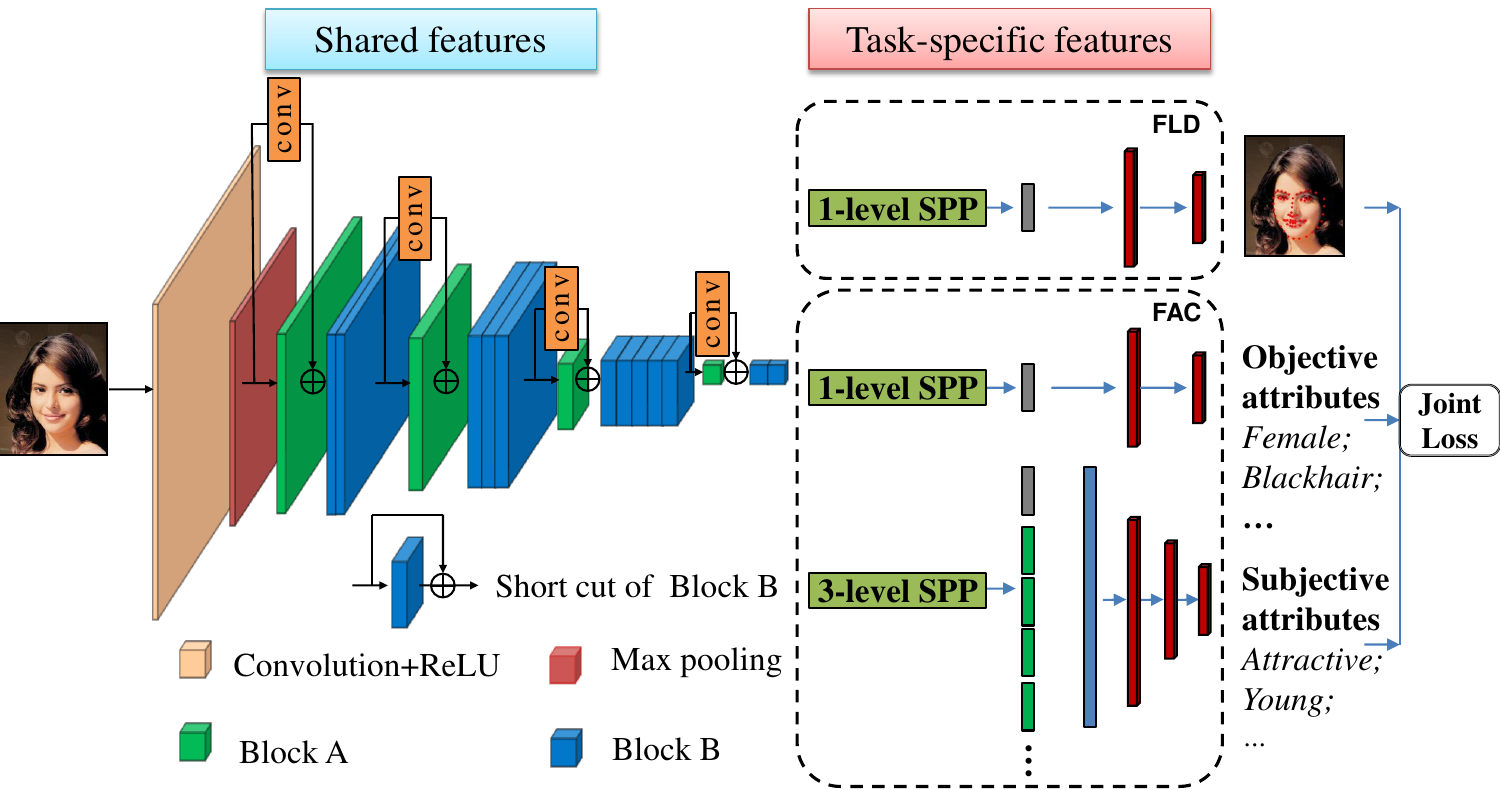}
\caption{{The framework of our proposed DMM-CNN method.  ResNet50 is used to extract the shared features and two sub-networks for FLD and FAC are jointly trained to extract task-specific features. Facial attributes are divided into objective attributes and subjective ones, where different network architectures are designed. Note that the shortcut of Block B is shown separately for clarity.}}
\label{fig:framework}
\end{figure*}

The overview of our proposed method is shown in Fig.~\ref{fig:framework}. In this paper, to extract the shared features,
we adopt ResNet50 \cite{b30} and remove the final global average pooling layer. Based on shared features, we further perform multi-task multi-label learning, where the task-specific features for two related tasks (FAC and FLD) are extracted.

Specifically, for the task of FAC, in order to deal with the diverse learning
complexities of facial attributes, we divide the facial attributes into two groups (objective attributes and subjective attributes) and design two different network architectures for these two groups (Section~\ref{sec:group}). In particular,
two different spatial pyramid pooling (SPP) layers, which extract different levels of semantic information, are respectively exploited for objective and subjective attributes in the network (Section~\ref{sec:spp}).
For the task of FLD, 72 facial landmark points are detected (Section~\ref{sec:fld}).
Hence, the whole network has three kinds of outputs (predicted outputs for objective attributes, subjective attributes and facial landmark regression).

During the training stage (Section~\ref{sec:train}),
the whole framework combines the losses from the two tasks into the final loss, where
 a novel adaptive weighting scheme is developed to automatically assign the loss weight to each facial attribute, such that the training concentrates on the classification of more difficult facial attributes.  Furthermore, to alleviate the problem of class imbalance, an adaptive thresholding strategy is developed to accurately predict the label of each attribute.
\subsection{CNN Architecture} \label{section}
In the following subsections, we respectively introduce the two groups of facial attributes, the SPP layer, and the task of facial landmark detection in detail.
\subsubsection{Objective Attributes and Subjective Attributes}
\label{sec:group}
To effectively exploit the intrinsic relationship and heterogeneity of facial attributes, the attributes can be divided into different groups \cite{b20, b29}. In this paper, we propose to classify facial attributes into two groups:  objective attributes (such as ``Attractive'', ``Big Nose'') and subjective attributes (such as ``Bald'', ``Male''). See Fig.~3 for more detail. Our design is based on the observation that state-of-the-art FAC methods often show much lower accuracy for predicting subjective attributes than objective attributes (for example, it is usually easier to classify the  ``Wearing Hat" and ``Wearing Eyeglasses" attributes than the ``Smiling" and ``Young" attributes). This is mainly because subjective attributes often appear in a subtle form, which makes the CNN model more difficult to learn the decision boundary.
In other words, objective and subjective attributes show different learning complexities.
Therefore, it is preferable to design different network architectures for these two groups of attributes.

In our implementation, the branch for learning the objective attributes consists of a 1-level SPP layer (see Section 3.2.2) and two fully connected layers with the output features of $1,024$ and $22$ (the number of objective attributes) dimensions, respectively. The branch for learning the subjective attributes consists of a 3-level SPP layer and  three fully connected layers with the  output features of $2,048$ , $1,024$ and $18$ (the number of subjective attributes) dimensions, respectively. In this manner, the network designed for the subjective attributes encodes higher-level semantic information (which is beneficial to predict the subjective attributes) than that designed for the objective attributes.

\subsubsection{The SPP Layer}
\label{sec:spp}
The Spatial Pyramid Pooling (SPP) layer proposed by He et al. \cite{b31} is introduced to deal with the problem of the fixed image size
 requirement for the CNN network. The SPP layer pools the features based on the top of the last convolutional layer and it is able to generate the fixed-length outputs regardless of the input size/scale. SPP aggregates the information from the deeper layer of the network, which effectively avoids the constraint for cropping or warping of the input image.

In this paper, we use the $1$-level SPP layer to extract features for objective attributes, and use the $3$-level SPP layer to extract features for subjective attributes (an $n$-level SPP layer divides a feature map into $n \times n$ blocks and then performs the max pooling operation in each block).
%
The size of the output feature maps for the $1$-level SPP layer and the $3$-level SPP layer are $2,048\times1$ and $28,672\times1$, respectively. Therefore, we can input the face images of any sizes to the networks by taking advantage of the SPP layers. As mentioned previously, the high-level semantic features are exploited to predict the subjective attributes, while the low-level appearance features are used to classify the objective attributes. The different levels of features are advantageous for classifying the two groups of attributes.


%
%

\subsubsection{Facial Landmark Detection (FLD)}
\label{sec:fld}

In this paper, two different but related tasks (i.e., FLD and FAC) are jointly trained by leveraging multi-task learning. Here, FAC is the target task while FLD is the auxiliary task. Under the paradigm of multi-task learning, the inherent dependencies
between the target task and the auxiliary task are exploited to effectively improve the
performance of FAC.
The landmark information of facial images is beneficial to improve the accuracy of FAC. For instance, the landmarks around the mouth can provide auxiliary information to help predict the ``smiling'' attribute.

Different from our previous work \cite{b18}, which  considers only 5 facial landmarks, we use the dlib library \footnote{http://dlib.net/} to obtain more facial feature points (72 facial landmarks in total) that outline the eyes, eyebrows, nose, mouth and facial boundary. Note that different facial attributes are usually related to different facial landmarks.
Therefore, using more facial landmarks is beneficial to improve the performance of FAC.  The FLD branch takes a 2,048 dimensional feature vector obtained by the 1-level SPP layer as the input and consists of two fully connected layers with the output features of $1,024$ and $144$ dimensions, respectively.


\subsection{Training}
\label{sec:train}
As we mention previously,  different facial attributes have different learning complexities. To deal with the diverse learning complexities of facial attributes, in addition to the adoption of different network architectures for objective and subjective attributes, we further propose
 a novel dynamic weighting scheme to automatically assign the loss weights to different attributes. Moreover, to alleviate the problem of class imbalance for multi-label training, an adaptive thresholding strategy is developed to predict the label of each attribute.

 In this paper, we use the mean square error (MSE) loss functions for simplicity in different tasks.
%

1) Facial landmark detection (FLD): The MSE loss for FLD is given as
\begin{equation}
\textit{L}_{FLD} = \frac{1}{N} \sum_{i=1}^{N}||\hat{\textit{\textbf{y}}}_{i}^{FLD} - \textit{\textbf{y}}_{i}^{FLD}||^{2},
\label{con:mse3}
\end{equation}
where $N$ is the number of training images. $\hat{\textit{\textbf{y}}}_{i}^{FLD} \in R^{2T}$ denotes the outputs (i.e., coordinate vector) of the facial landmarks ($T$ is the number of facial landmarks, and we use 72 facial landmarks in this paper) obtained from the network.
$\textit{\textbf{y}}^{FLD}\in R^{2T}$ represents the ground-truth coordinate vector.

2) Facial attribute classification: The MSE loss for FAC is given as
\begin{equation}
\textit{L}_{FAC}^{j} = \frac{1}{N} \sum_{i=1}^{N}(\hat{\textit{y}}_{i,j}^{FAC} - \textit{y}_{i,j}^{FAC})^{2},
\label{con:mse1}
\end{equation}
where $\hat{\textit{y}}^{FAC}_{i,j}$ and $\textit{y}^{FAC}_{i,j}$ ($\in \{1,-1\}$) represent the predicted output and the label corresponding to the $j$-th attribute of the $i$-th training image, respectively.

3) The joint loss: The joint loss consists of the losses for FLD and FAC, which can be written as
\begin{equation}
  L = \sum_{j=1}^{J} {{\lambda}_{t}^j}{L}_{FAC}^{j}+ \beta \textit{L}_{FLD},
 \label{con:jointloss}
\end{equation}
where $J$ is the total number of facial attributes. $\boldsymbol{\lambda}_t = [\lambda^1_t, \lambda^2_t, \cdots, \lambda^J_t]^T$ represents the weight vector corresponding to the $J$ facial attributes during the $t$-th iteration. $\beta$ is the regularization parameter (we empirically set $\beta$ to 0.5).

4) Dynamic weighting scheme. In this paper, we propose a dynamic weighting scheme to automatically assign weights to all facial attributes. The loss weights are dynamically assigned according to the validation loss trend \cite{b27}.
 Specifically, the weights are defined as
\begin{equation}
{\lambda}_t^{j} = \left| \frac{{{L}_{FAC}^{j,VAL}}(t) -{{L}_{FAC}^{j,VAL}}(t-1) }{{{L}_{FAC}^{j,VAL}}(t-1)} \right|,
\label{con:lambda}
\end{equation}
where ${{L}_{FAC}^{j,VAL}}(t)$ is the validation loss (computed according to Eq.~(\ref{con:mse1}) for each attribute on the validation set) during the $t$-th iteration of the training. In this way, the weights corresponding to the facial attributes will be assigned low values if the validation loss does not decrease, while those will be given high values if the validation loss significantly drops.

During the initial training process, the easily-classified attributes are assigned large weights so that their corresponding MSE losses can be quickly reduced. As the iteration proceeds, the MSE losses for the hardly-classified attributes become relatively larger and  drop slowly, while those for the easily-classified ones become smaller. Therefore, in the later stage of the training process, the network focuses on the training of classification of the hardly-classified attributes (note that the loss for each attribute is composed of the multiplication of the weight and its corresponding MSE loss).


Note that the weighting schemes are also developed in \cite{Liu} and \cite{b32}. However, the differences between the proposed dynamic weighting scheme and those in \cite{Liu, b32} are significant. In \cite{Liu}, the weights are computed based on the rate of training
loss changes. In  \cite{b32}, the weights are computed according
to the validation loss and the mean validation loss trend.
Note that, the validation loss may not be appropriate for
determining the weight. In contrast, the proposed dynamic weighting
scheme computes the weights only based on the validation
loss trend. Moreover, the weights in \cite{Liu} are obtained according to the average training loss (in the training set) in each epoch
over several iterations. Different from \cite{Liu}, the weighting
scheme in  \cite{b32} and our proposed one take advantage of
the validation set, which can be beneficial to improve the
generalization ability of a learned model (since the validation set is not directly used to compute gradients during
the back-propagation process). In  \cite{b32}, the validation loss
is computed on a small batch (containing only 10 validation images)
during each iteration, while it is computed on the whole
validation set for every $P$ iterations in our method. Therefore, the proposed dynamic weighting scheme shows more stable loss reduction.

5) Adaptive thresholding strategy. We predict the label of the $j$-th facial attribute $\hat{l}_j$ according to the final output of the network:
\begin{equation}
\hat{l}_j=\left\{
\begin{aligned}
 1 & , & output > \tau_j  \\
-1 & , & output \leq \tau_j
\end{aligned}
\right.,
\label{con:score}
\end{equation}
where $\tau_j$ is the threshold parameter.
If the predicted output is larger than the threshold $\tau_j$, a positive label is assigned.

Existing FAC methods usually set the threshold $\tau_j$ to be 0. However, due to the problem of class imbalance (i.e., the number of samples for one class is significantly larger than that for the other class for one attribute), using the fixed threshold is not an optimal solution, especially for some highly imbalanced facial attributes. In this paper, we introduce an adaptive thresholding strategy, which adaptively updates the threshold as follows: 

\begin{equation}
{\boldsymbol{\tau}_{t} = \boldsymbol{\tau}_{t-1} + \gamma l(\textbf{N}^{FP}_{t}-\textbf{N}^{FN}_{t})/V}
\label{con:t}
\end{equation}
where $\boldsymbol{\tau}_{t} \in R^{J}$ is the threshold for the $t$-th iteration. $V$ is the number of samples in the validation set. $l$ is the current epoch. $\textbf{N}^{FP}_{t}\in R^{J}$ ($\textbf{N}^{FN}_{t}\in R^{J}$) represents the number of false positive (false negative) in the validation set for the $t$-th iteration. The larger the value of $\textbf{N}^{FP}_{t}$ is (or the smaller the value of $\textbf{N}^{FN}_{t}$ is), the higher the value of the threshold should be. Hence, the difference between $\textbf{N}^{FP}_{t}$  and $\textbf{N}^{FN}_{t}$ can be used to adjust the threshold. We also consider the current epoch in Eq.~(\ref{con:t}), since more attention should be paid to false predictions as the training epoch increases (the threshold is adapted to a larger value). $\gamma$ is the fixed parameter (we experimentally set it to $0.01$ in this paper).

The training stage of the proposed DMM-CNN method is summarized in Algorithm 1.

\begin{algorithm}[!t]
\caption{ The training stage of the proposed DMM-CNN method.}
\begin{algorithmic}[1]
\REQUIRE
 Training data and validation data. Initialized parameters $\boldsymbol{\theta}$ of CNN. The maximum number of iterations $M$. The updating interval $P$.
\ENSURE
The model parameters $\boldsymbol{\theta}$ of the trained CNN model.
\STATE $loop=0$, $t=1$;
\WHILE {$loop \leq M$}
\IF{$loop \% P=0$}
\STATE Calculate the validation loss of facial attributes according to Eq.~(\ref{con:mse1});

\STATE Update $\boldsymbol{\tau}_t$ according to Eq.~(\ref{con:t});
\STATE Update $\boldsymbol{\lambda}_t$ according to Eq.~(\ref{con:lambda});
\STATE $t = t + 1$;
\ENDIF
\STATE Calculate the joint loss $L$ according to Eq.~(\ref{con:jointloss});
\STATE Update the parameters $\boldsymbol{\theta}$ using the stochastic gradient descent technique;
\STATE $loop = loop + 1$;
\ENDWHILE
\label{code:recentEnd}
\end{algorithmic}
\end{algorithm}
%
%

\section{Experiments}
In this section, we firstly introduce two public FAC datasets used for evaluation.
Then, we perform an ablation study to discuss the influence of every component of the proposed DMM-CNN method.
Finally, we compare the proposed DMM-CNN method with several state-of-the-art FAC methods.

\subsection{Datasets and Parameter Settings}
CelebA \cite{b33} is a large-scale face dataset, which is provided with the labeled bounding box and the annotations of 5 landmarks and 40 facial attributes. It contains 162,770 images for training, 19,867 images for validation and  19,962 images for testing. The images in CelebA cover large pose variations and background clutter. LFWA \cite{b34} is another challenging face dataset that contains 13,143 images with 73 binary facial attribute annotations. We select the same 40 attributes from LFWA as CelebA. For LFWA, we fine-tune the model trained on CelebA and use both the original and the deep funneled images of LFWA as the training set to prevent over-fitting. As a result, 13,144 images are used for training and 6,571 images for testing for LFWA.
Since LFWA does not provide the validation set, we directly update the dynamic weights and use the adaptive thresholding strategy on the training set.

The proposed method is implemented based on the open source deep learning framework pytorch, where one NVIDIA TITAN X GPU is used to train the model for 15 epochs with the batch size of 64. The base learning rate is set to 0.001 and we multiply the learning rate by 0.1 when the validation loss stops decreasing.
The model size is about 360M.


\subsection{Ablation Study}
In this subsection, we will give an ablation study to evaluate the effectiveness of different components of the proposed DMM-CNN on the CelebA and LFWA datasets.

\begin{figure*}[!t]
\centerline{\includegraphics[height=49mm,width=1\linewidth]{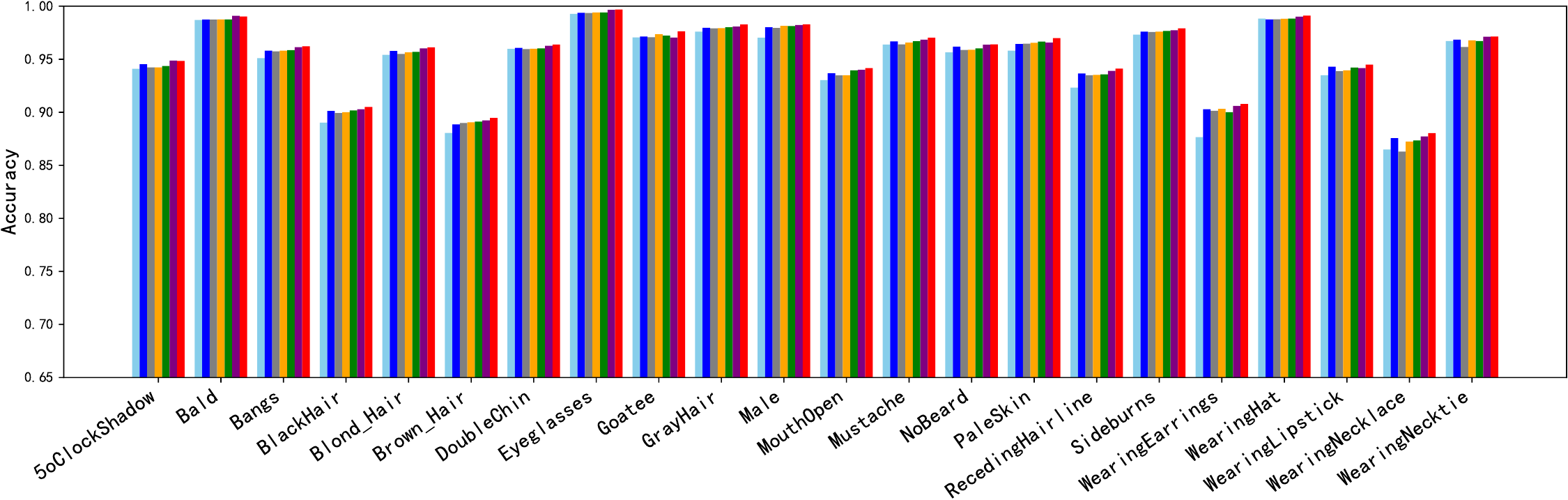}}
\centerline{\includegraphics[height=49mm,width=1\linewidth]{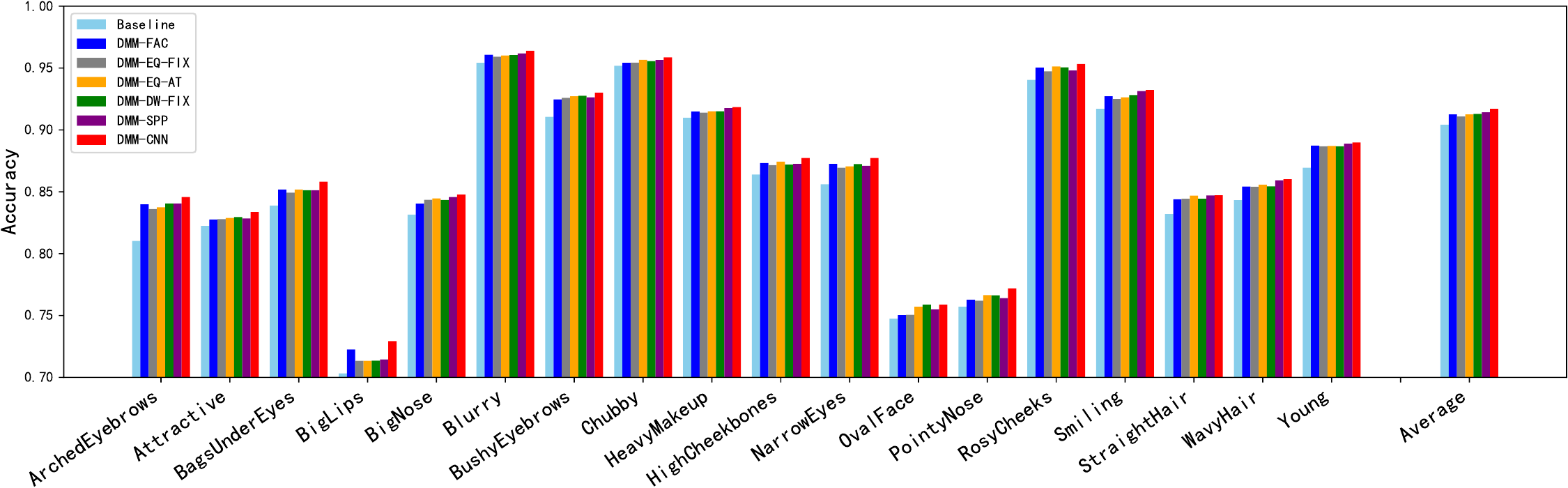}}
\caption{Performance comparison between different variants of the proposed DMM-CNN method on the CelebA dataset, where the upper panel shows the objective attributes, while the lower panel shows the subjective attributes.}
\label{fig:ablation}
\end{figure*}

\begin{figure*}[!t]
\centerline{\includegraphics[height=49mm,width=1\linewidth]{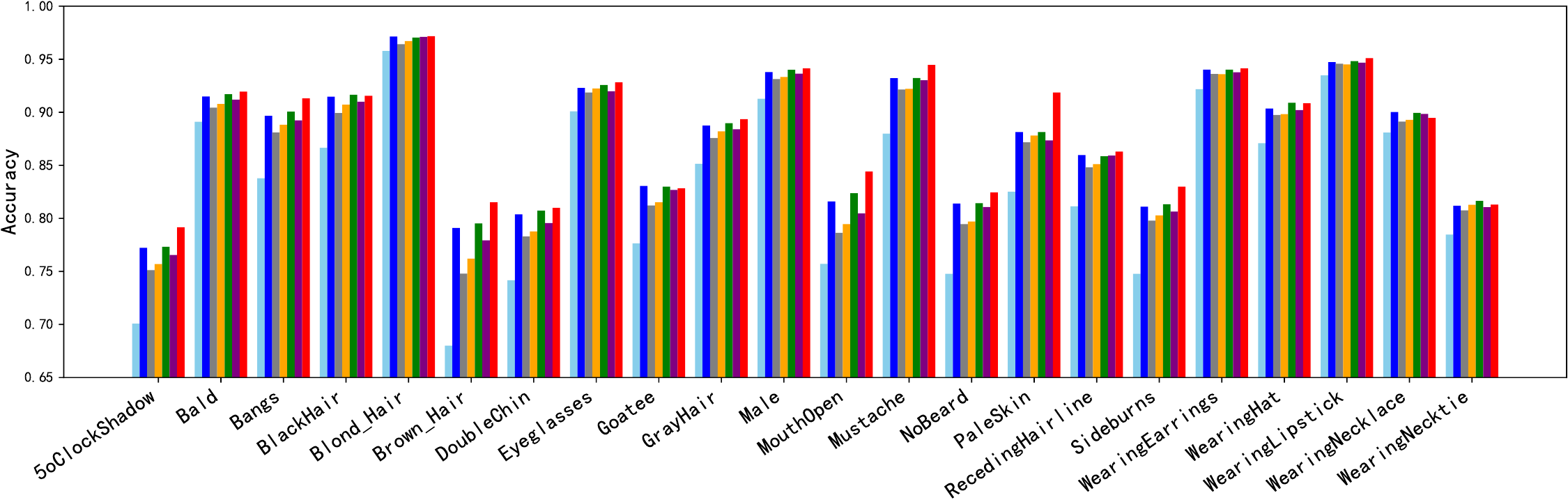}}
\centerline{\includegraphics[height=49mm,width=1\linewidth]{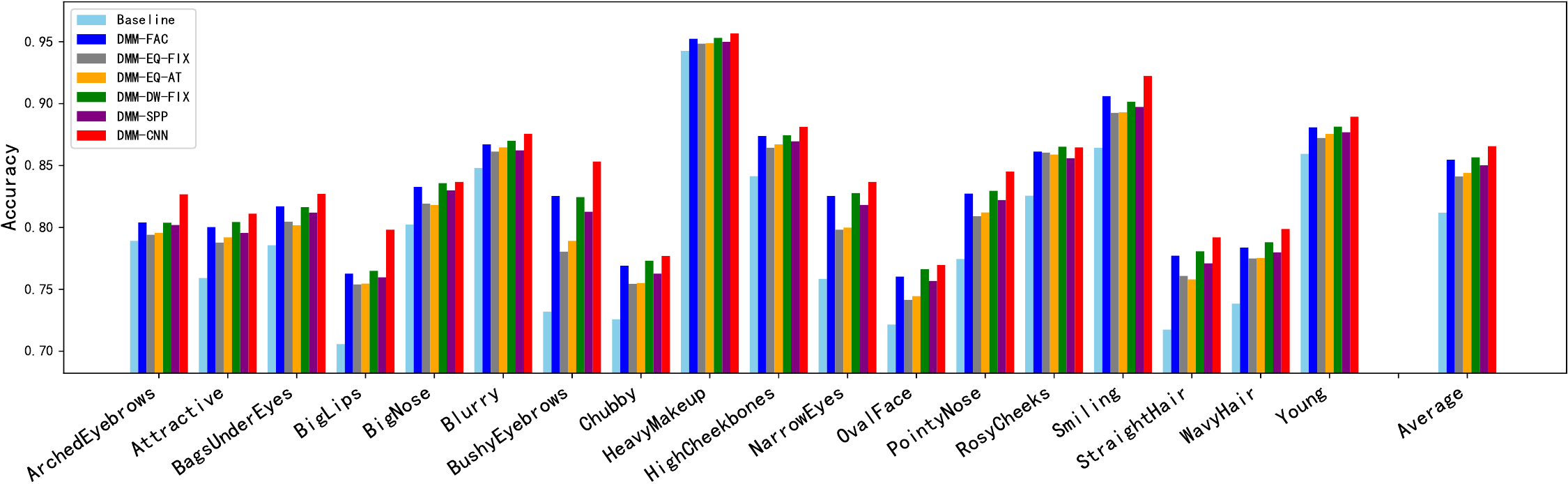}}
\caption{Performance comparison between different variants of the proposed DMM-CNN method on the LFWA dataset, where the upper panel shows the objective attributes, while the lower panel shows the subjective attributes.}
\label{fig:ablation2}
\end{figure*}

\begin{table}[!t]
	\centering
	\caption{The details of the seven variants. FLD denotes facial landmark detection. DW denotes dynamic weights. AT denotes the adaptive thresholding.  AG denotes attribute grouping.}
	\renewcommand\arraystretch{1.2}
	\small
	\setlength{\tabcolsep}{0.8mm}{
	\begin{tabular}{ccccc}
		\toprule  
		Variants & FLD & DW & AT &  AG \\
		\midrule
		Baseline&    &  &   & \\
		DMM-FAC&    & \checkmark &   \checkmark & \checkmark \\
		DMM-EQ-FIX&  \checkmark  &  &   & \checkmark \\
		DMM-EQ-AT&  \checkmark  &  &   \checkmark & \checkmark \\
	    DMM-DW-FIX&  \checkmark  & \checkmark &   & \checkmark \\
	    DMM-SPP&  \checkmark  & \checkmark &   \checkmark & \\
	    DMM-CNN&  \checkmark  & \checkmark &   \checkmark & \checkmark \\

		\bottomrule  
	\end{tabular}}
	\label{tab:dividing}
\end{table}

We evaluate several variants of the proposed DMM-CNN method. Specifically, Baseline represents that we only use ResNet50 (with 40 output units) to extract features and classify the attributes.  DMM-FAC represents that we only perform the single task of FAC without using the auxiliary task of FLD. DMM-EQ-FIX represents that we use  equal loss weights (i.e., 1.0) for all the attributes without relying on the proposed dynamic weighting scheme, and the fixed threshold (i.e., 0.0) to predict the label of each attribute instead of using the adaptive threshold.  DMM-EQ-AT represents that we use equal loss weights for all the attributes and the proposed adaptive thresholding strategy. DMM-DW-FIX represents that we use the dynamic weighting scheme and the fixed threshold.  DMM-SPP represents that we use the 3-level SPP layer and three fully connected layers to predict all the attributes (using the same network architecture as the subjective attributes branch) without attribute grouping. DMM-CNN is the proposed method. The details of all the competing variants are listed in Table 1.

The performance (i.e., the accuracy rate) obtained by different variants is shown in Fig.~\ref{fig:ablation}. We have the following observations:
\begin{itemize}
\item Compared with the Baseline, all the other variants achieve better performance (especially on the  ``ArchedEyebrowns", ``Big Lips" and ``Narrow Eyes" attributes), which demonstrates the importance of using task-specific features for FAC.
\item By comparing DMM-FAC with DMM-CNN, we can see that multi-task learning is beneficial to improve the performance of FAC by exploiting the intrinsic relationship between FAC and FLD.
\item DMM-DW-FIX achieves higher classification accuracy compared with DMM-EQ-FIX in terms of average classification rate, which shows the superiority of using the dynamic weighting scheme.
\item   The average classification rate obtained by DMM-EQ-AT is higher than that obtained by DMM-EQ-FIX, which shows the effectiveness of using the adaptive thresholding strategy.
    \item Compared with the baseline, the improvements of DMM-DW-FIX and  DMM-EQ-AT on LFWA are more evident than those on CelebA.  Specifically, DMM-DW-FIX achieves 5.52\% (0.91\%) improvement in accuracy, while DMM-EQ-AT obtains 3.98\% (0.95\%) improvement in accuracy on LFWA (CelebA).  The improvements on CelebA are marginal. Such a phenomenon is also observed in some papers \cite{He2018,Huang,Lu2017}.  This may be because that the discrepancy between the distributions from the training set and the test set of CelebA is large, and these exists some noise in the CelebA labels especially for the subjective attributes \cite{Hand2018}, leading to the difficulty of significant improvements in the test set of CelebA.
    \item Compared with DMM-SPP, DMM-CNN achieves better accuracy (i.e., 0.30\% and 1.81\% improvements on CelebA and LFWA, respectively).  Therefore, designing different network architectures, which take into account the diverse learning complexities of facial attributes, is beneficial to improve the performance of FAC.
        \item Among all the variants, DMM-CNN achieves the best accuracy, which can be attributed to the multi-task learning and multi-label learning framework that exploits the different learning complexities of facial attributes.
\end{itemize}

The loss weighting scheme plays a critical role in the performance of FAC.  we compare the performance of different weighting schemes. Specifically, we evaluate the following four representative weighting schemes: 1) Uniform Weighting (UW) scheme, where all the weights corresponding to different attributes are set to 1.0; 2) Dynamic Weight Average (DWA) scheme proposed in \cite{Liu}, where the rate of loss change in the training set is used to automatically learn the weights;
3) Adaptive Weighting (AW) scheme proposed in \cite{b32}, where both the validation loss and the mean validation loss trend in a batch are used to obtain the weights; 4) The proposed dynamic weighting scheme, which takes advantage of the rate of validation loss changes in the whole validation set. Table 2 gives the experimental results of different weighting schemes on the CelebA and LFWA datasets. We can see that our method with the proposed dynamic weighting scheme achieves the best performance compared with other weighting schemes, which can validate the effectiveness of the proposed one.

\begin{table}[!t]
	\centering
	\caption{Experimental results of different weighting schemes on the CelebA and LFWA datasets.}
	\renewcommand\arraystretch{1.2}
	\small
	\setlength{\tabcolsep}{0.8mm}{
	\begin{tabular}{ccc}
		\toprule  
		\multirow{2}{*}{Weighting Scheme}
		& \multicolumn{2}{c}{Mean accuracy (\%)} \\
	    \cline{2-3}
		& CelebA & LFWA \\
		\midrule
		UW &  91.08  & 84.11 \\
		DWA~\cite{Liu} &  91.36  & 84.78\\
		AW~\cite{b32} &  91.65  & 85.02\\
		Our proposed scheme & 91.70 & 86.56 \\
	
		\bottomrule  
	\end{tabular}}
	\label{tab:dynamicweights}
\end{table}

\begin{figure}[!t]
\centerline{\includegraphics[width=0.85\linewidth]{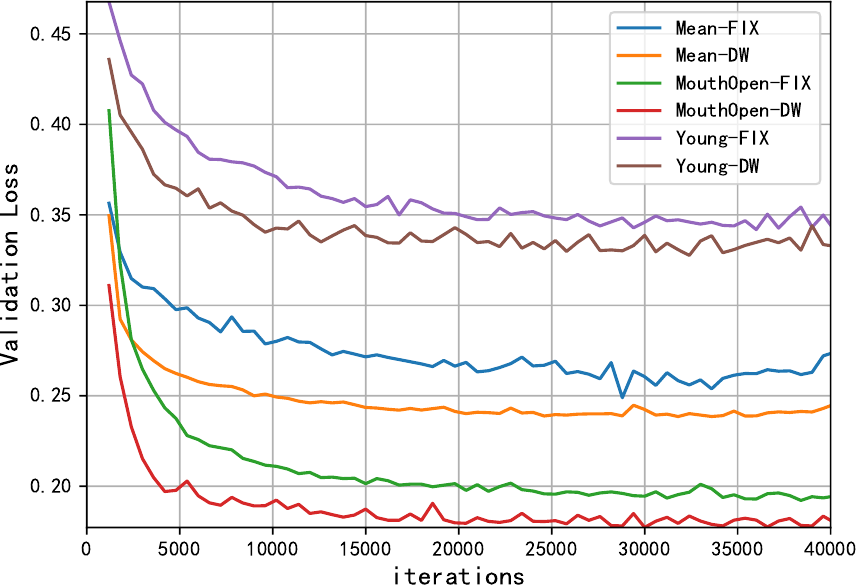}}
\caption{Changes of the validation loss with the number of iterations using the proposed dynamic weighting scheme and the fixed weighting scheme during the training stage. Here, Mean-FIX, MouthOpen-FIX, Young-FIX, Mean-DW,  MouthOpen-DW and Young-DW denote the mean validation loss, two attribute losses using the fixed weighting scheme and the dynamic weighting scheme, respectively.}
\label{fig:validationloss}
\end{figure}

In Fig.~\ref{fig:validationloss}, we further visualize the changes of mean validation loss and two representative attribute losses (i.e., for the objective attribute ``MouthOpen'' and the subjective attribute ``Young'') on the validation set during the training stage. Here, the proposed
dynamic weighting scheme and the fixed weighting scheme (i.e., the weight is set to 1.0 for each attribute) are respectively employed. We can observe that the mean validation loss based on the dynamic weighting scheme decreases faster than that based on the fixed weighting scheme.
The training of the objective attribute (i.e., ``MouthOpen'') converges much faster than the subjective attribute (i.e., ``Young''). During the initial training stage, the loss of the ``MouthOpen'' attribute quickly drops and converges after about 15,000 iterations.
In contrast, the loss of  the ``Young'' attribute slowly drops and converges after about  30,000 iterations.
As the training proceeds, the network focuses on classifying those difficult subjective attributes.
 In general, the loss using the dynamic weighting scheme usually drops more and faster than that using the fixed weighting scheme.
This reveals that dynamic weights are of vital importance when optimizing the multi-label learning task having different learning complexities.

\begin{figure}[!t]
\centerline{\includegraphics[width=0.85\linewidth]{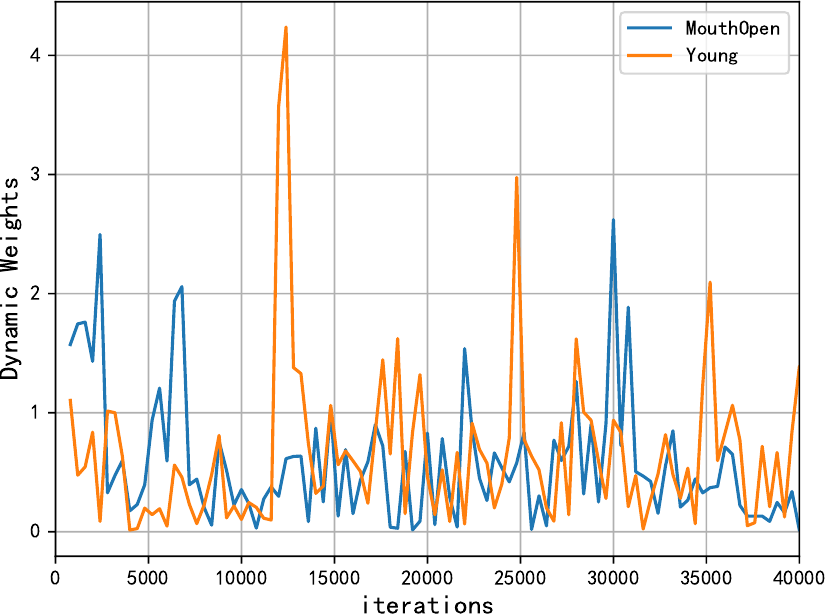}}
\caption{Curves of dynamic weights during the training stage.}
\label{fig:weights}
\end{figure}

\begin{figure}[!t]
\centerline{\includegraphics[width=0.8\linewidth]{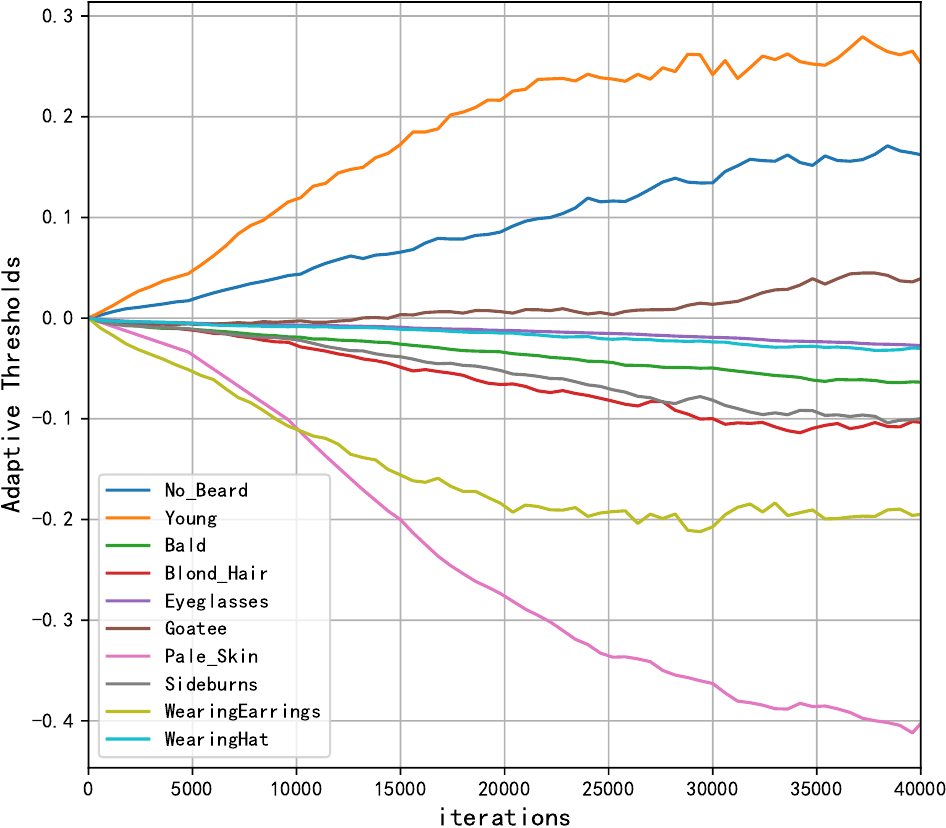}}
\caption{Curves of adaptive thresholds during the training stage.}
\label{fig:adaptivethresholding}
\end{figure}

We visualize the changes of dynamic weights and adaptive threshold in the training stage in  Fig.~\ref{fig:weights} and  Fig.~\ref{fig:adaptivethresholding}, respectively.


Firstly, in Fig.~\ref{fig:weights}, the curves of two dynamic weights corresponding to two representative facial attributes (i.e., ``MouthOpen'' and ``Young'') during the training stage are given.
We can observe that the changes of the dynamic weights corresponding to the two attributes are unstable. This is mainly because the proposed weighting scheme dynamically assigns the weight to each attribute according to the rate of the attribute loss changes (see Eq.~(4)). In other words,
when the loss of an attribute significantly drops, a large weight will be assigned to this attribute (since the learning process of this attribute does not converge). Therefore,  the dynamic weights reflect the learning rates of different attributes, which may significantly vary.
However, note that the losses of these two attributes keep decreasing and converge stably (see Fig.~5).

Secondly, in Fig.~\ref{fig:adaptivethresholding}, the curves of adaptive thresholds corresponding to the ten randomly-chosen facial attributes during  the training stage are given.
We can observe that the changes of thresholds are stable. This is mainly due to the fact that
the difference between the number of false positive and that of false negative is used to adjust the threshold. As the iteration goes, the difference becomes more stable.

%

%

\begin{table*}[htbp]
	\setlength{\abovecaptionskip}{0.cm}
	\setlength{\belowcaptionskip}{-0.cm}
	\centering
	\caption{ The classification accuracy (\%) obtained by all the competing methods on the CelebA and LFWA datasets. ``-'' indicates that the corresponding results are not provided by the method. The accuracy for each attribute obtained by the proposed method is highlighted in bold.}

	\scriptsize
	\setlength{\tabcolsep}{0.8mm}{
	\begin{tabular}{c|c|c|c|c|c|c|c|c|c|c|c|c|c|c|c|c|c|c|c|c|c|c}
		\toprule  
		&& \rotatebox{90}{5 o'clock Shadow}	&	\rotatebox{90}{Arched Eyebrows}	&	\rotatebox{90}{Attractive}	&	\rotatebox{90}{Bags Under Eyes}	&	\rotatebox{90}{Bald}	&	\rotatebox{90}{Bangs}	&	
		\rotatebox{90}{Big Lips}	&	\rotatebox{90}{Big Nose}	&	\rotatebox{90}{Black Hair}	&	\rotatebox{90}{Blond Hair}	&	\rotatebox{90}{Blurry}	&	\rotatebox{90}{Brown Hair}	&	\rotatebox{90}{Bushy Eyebrows}	&	\rotatebox{90}{Chubby}	&	\rotatebox{90}{Double Chin}	&	\rotatebox{90}{Eyeglasses}	&	\rotatebox{90}{Goatee}	&	\rotatebox{90}{GrayHair}	&	\rotatebox{90}{Heavy Makeup}	&	\rotatebox{90}{High Cheekbones}	&	\rotatebox{90}{Male} \\
		\midrule
		 \multirow{7}{*}{CelebA}
		 & PANDA & 88.00 & 78.00	& 81.00& 	79.00	& 96.00	& 92.00& 	67.00& 	75.00& 	85.00& 	93.00& 	86.00& 	77.00& 	86.00& 	86.00& 	88.00& 	98.00& 	93.00& 	94.00& 	90.00& 	86.00& 	97.00

		 \\
		 & LNets+ANet&	91.00	&	79.00	&	81.00	&	79.00	&	98.00	&	95.00	&	68.00	&	78.00	&	88.00	&	95.00	&	84.00	&	80.00	&	90.00	&	91.00	&	92.00	&	99.00	&	95.00	&	97.00	&	90.00	&	88.00	&	98.00 \\
		 & MOON	&	94.03	&	82.26	&	81.67	&	84.92	&	98.77	&	95.80	&	71.48	&	84.00	&	89.40	&	95.86	&	95.67	&	89.38	&	92.62	&	95.44	&	96.32	&	99.47	&	97.04	&	98.10	&	90.99	&	87.01	&	98.10																				
		  \\
		
		  & NSA & 93.13	&	82.56	&	82.76	&	84.86	&	98.03	&	95.71	&	69.28	&	83.81	&	89.03	&	95.76	&	95.96	&	88.25	&	92.66	&	94.94	&	95.80	&	99.51	&	96.68	&	97.45	&	91.59	&	87.61	&	97.95
		  \\
		
		 & MCNN-AUX	&	94.51	&	83.42	&	83.06	&	84.92	&	98.90	&	96.05	&	71.47	&	84.53	&	89.78	&	96.01	&	96.17	&	89.15	&	92.84	&	95.67	&	96.32	&	99.63	&	97.24	&	98.20	&	91.55	&	87.58	&	98.17 \\																			
        & MCFA & 94.00	&	83.00	&	83.00	&	85.00	&	99.00	&	96.00	&	72.00	&	84.00	&	89.00	&	96.00	&	96.00	&	88.00	&	92.00	&	96.00	&	96.00	&	100.00	&	97.00	&	98.00	&	92.00	&	87.00	&	98.00
        \\
        & {GNAS} & 94.76&	84.25&	83.06&	85.87&	98.96&	96.20&	71.79&	85.10&	90.24&	96.11&	96.42&	89.75&	92.99&	95.93&	96.48&	99.69&	97.59&	98.37&	91.82&	88.05&	98.50
        \\
        &{AW-CNN} & -&	 -&	 -&	 -&	 -&	 -&	 -&	 -&	 -&	 -&	 -&	 -&	 -&	-&	-&	 -&	 -&	 -&	 -&	 -&	 -
        \\
        &{PS-MCNN-LC} & 96.60	& 85.77&	84.39&	87.29&	99.41&	98.00&	73.13&	86.40&	91.66&	97.93&	98.00&	91.03&	94.51&	97.66&	98.29&	99.85&	97.74&	98.66&	93.31&	89.50&	98.81
        \\
         																	
		 & DMM-CNN	&	\textbf{94.84}	&	\textbf{84.57}	&	\textbf{83.37}	&	\textbf{85.81}	&	\textbf{99.03}	&	\textbf{96.22}	&	\textbf{72.93}	&	\textbf{84.78}	&	\textbf{90.50}	&	\textbf{96.13}	&	\textbf{96.40}	&	\textbf{89.46}	&	\textbf{93.01}	&	\textbf{95.86}	&	\textbf{96.39}	&\textbf{99.69}	&	\textbf{97.63}	&	\textbf{98.27}	& 	\textbf{91.85}	&		\textbf{87.73}	&	\textbf{98.29} \\																				
		 \midrule
		 \midrule	
		 \multirow{7}{*}{LFWA}
		 & PANDA & 84.00 &	79.00&	81.00&	80.00&	84.00&	84.00&	73.00&	79.00&	87.00&	94.00&	74.00&	74.00&	79.00&	69.00&	75.00&	89.00&	75.00&	81.00&	93.00&	86.00&	92.00
		 \\

		 & LNets+ANet	&	84.00	&	82.00	&	83.00	&	83.00	&	88.00	&	88.00	&	75.00	&	81.00	&	90.00	&	97.00	&	74.00	&	77.00	&	82.00	&	73.00	&	78.00	&95.00	&	78.00	&	84.00	&	95.00	&	88.00	&	94.00 \\																			
		 & NSA & 77.59	&	81.72	&	80.16	&	82.62	&	91.88	&	90.71	&	78.97	&	83.13	&	92.49	&97.47	&	86.42	&	80.93	&	84.26	&	76.06	&	80.49	&	91.50	&	83.01	&	88.46	&	95.39	&	88.34	&	92.60

		 \\																			
		 & MCNN-AUX	&	77.06	&	81.78	&	80.31	&	83.48	&	91.94	&	90.08	&	79.24	&	84.98	&92.63	&	97.41	&	85.23	&	80.85	&	84.97	&	76.86	&81.52	&	91.30	&	82.97	&	88.93	&	95.85	&88.38	&	94.02 \\																				
& MCFA & 75.00	& 79.00 &	77.00 &	79.00 &	91.00 &	89.00 &	75.00 &	81.00 &	91.00 &	97.00 &	86.00	& 77.00	& 76.00	& 74.00 & 77.00	& 91.00	& 80.00 &	88.00 & 94.00 &	85.00 & 93.00
\\
 & {GNAS} & -&	-&	-&	-&	-&	-&	-&	-&	-&	-&	-&	-&	-&	-&	-&	-&	-&	-& -&	-&	-
 \\		
 		
   &{PS-MCNN-LC} & 78.17&	83.53&	81.84&	86.74&	92.60&	91.45&	82.70&	86.48&	92.96&	98.51&	87.20&	81.87&	85.72&	78.11&	86.70&	92.78&	84.11&	91.04& 96.60&	88.77&	95.18
   \\																
		 & DMM-CNN	&	\textbf{79.18}	&	\textbf{82.70}	&	\textbf{81.10}	&	\textbf{82.70}	&	\textbf{91.96}	&	\textbf{91.30}	&	\textbf{79.82}	&	\textbf{83.67}	&	\textbf{91.55}	&	\textbf{97.17}	&	\textbf{87.58}	&	\textbf{81.56}	&	\textbf{85.33}	&	\textbf{77.66}	&	\textbf{80.98}	&	\textbf{92.83}	&	\textbf{82.82}	&	\textbf{89.38}	&	\textbf{95.68}	&	\textbf{88.13}	&	\textbf{94.14} \\																				

		\midrule  
	
	    &&	\rotatebox{90}{MouthOpen}	&	\rotatebox{90}{Mustache}	&	\rotatebox{90}{NarrowEyes}	&	\rotatebox{90}{NoBeard}	&	\rotatebox{90}{OvalFace}	&	\rotatebox{90}{PaleSkin}	&	\rotatebox{90}{PointyNose}	&	\rotatebox{90}{RecedingHairline}	&	\rotatebox{90}{RosyCheeks}	&	\rotatebox{90}{Sideburns}	&	\rotatebox{90}{Smiling}	&	\rotatebox{90}{Straight Hair}	&	\rotatebox{90}{WavyHair}	&	\rotatebox{90}{WearingEarrings}	&	\rotatebox{90}{WearingHat}	&	\rotatebox{90}{WearingLipstick}	&	\rotatebox{90}{WearingNecklace}	&	\rotatebox{90}{WearingNecktie}	&	\rotatebox{90}{Young}	&		&	\rotatebox{90}{Average} \\
	    \midrule
	
	    \multirow{7}{*}{CelebA}&
	    PANDA & 	93.00& 	93.00& 	84.00& 	93.00& 	65.00& 	91.00& 	71.00& 	85.00& 	87.00& 	93.00& 	92.00& 	69.00& 	77.00& 	78.00& 	96.00& 	93.00& 	67.00& 	91.00& 	84.00& 	& 85.43
	    \\
	     & LNets+ANet	&	92.00	&	95.00	&	81.00	&	95.00	&	66.00	&	91.00	&	72.00	&	89.00	&	90.00	&	96.00	&	92.00	&	73.00	&	80.00	&	82.00	&	99.00	&	93.00	&	71.00	&	93.00	&	87.00	&		&	87.33 \\																				
	    & MOON	&	93.54	&	96.82	&	86.52	&	95.58	&	75.73	&	97.00	&	76.46	&	93.56	&	94.82	&	97.59	&	92.60	&	82.26	&	82.47	&	89.60	&	98.95	&	93.93	&	87.04	&	96.63	&	88.08	&		&	90.94
	    \\
	    & NSA & 93.78	&	95.86	&	86.88	&	96.17	&	74.93	&	97.00	&	76.47	&	92.25	&	94.79	&	97.17	&	92.70	&	80.41	&	81.70	&	89.44	&	98.74	&	93.21	&	85.61	&	96.05	&	88.01	&	& 90.61
	
	    	\\																			
	    & MCNN-AUX	&	93.74	&	96.88	&	87.23	&	96.05	&	75.84	&	{97.05}	&	{77.47}	&	93.81	&	95.16	&	97.85	&	92.73	&	83.58	&	83.91	&	90.43	&	99.05	&	94.11	&	86.63	&	96.51	&	88.48	&		&	91.29	\\																			

		 & MCFA  & 93.00	&	97.00	&	87.00	&	96.00	&	75.00	&	97.00	&	77.00	&	94.00	&	95.00	&	{98.00}	&	93.00	&	{85.00}	&	85.00	&	90.00	&	99.00	&	94.00	&	88.00	&	97.00	&	88.00	&   & 91.23
		 \\
	    		  &{GNAS}& 94.16	& 97.03&	87.66&	96.3&	75.57&	97.24&	78.24&	93.94&	95.01&	97.96&	93.24&	84.77&	84.52&	90.98&	99.12&	94.41&	87.61&	96.76&	88.89&	&91.63
\\
&{AW-CNN} & -	& -&	-&	-&	-&	-&	-&	-&	-&	-&	-&	-&	-&	-&	-&	-&	-&-&-&	&91.80
\\
&  {PS-MCNN-LC} & 95.99&	98.56&	89.07&	98.03&	77.43&	98.84&	79.32&	95.85&	96.92&	98.22&	94.85&	85.96&	86.39&	92.66&	99.43&	95.70&	88.98&	98.52&	90.54&	& 92.98
\\								
	    & DMM-CNN	&	\textbf{94.16}	&	\textbf{97.03}	&	\textbf{87.73}	&	\textbf{96.41}	&	\textbf{75.89}	&	\textbf{97.00}	&	\textbf{77.19}	&	\textbf{94.12}	&	\textbf{95.32}	&	\textbf{97.91}	&	\textbf{93.22}	&	\textbf{84.72}	&	\textbf{86.01}	&	\textbf{90.78}	&	\textbf{99.12}	&	\textbf{94.49}	&	\textbf{88.03}	&	\textbf{97.15}	&	\textbf{88.98}	&		&	\textbf{91.70}	\\	
	    \midrule
	    \midrule																		
	    																				
	    \multirow{7}{*}{LFWA}
	    & PANDA &	78.00&	87.00&	73.00&	75.00&	72.00&	84.00&	76.00&	84.00&	73.00&	76.00&	89.00&	73.00&	75.00&	92.00&	82.00&	93.00&	86.00&	79.00&	82.00&	& 81.03
	    \\
	    & LNets+ANet	&	82.00	&	92.00	&	81.00	&	79.00	&	74.00	&	84.00	&	80.00	&	85.00	&	78.00	&	77.00	&	91.00	&	76.00	&	76.00	&	94.00	&	88.00	&	95.00	&	88.00	&	79.00	&	86.00	&		&	83.85		\\																		
	
	    & NSA  & 82.50	&	92.97	&	82.75	&	80.77	&	76.80	&	90.97	&	84.20	&	84.90	&	87.08	&	81.76	&	90.80	&	78.91	&	78.28	&	94.75	&	90.23	&	94.07	&	89.59	&	81.40	&	85.68	&		&	85.82
	
	    \\																	
	    & MCNN-AUX	&	83.51	&	93.43	&	82.86	&	82.15	&	{77.39}	&	{93.32}	&	84.14	&	86.25	&	{87.92}	&	{83.13}	&	91.83	&	78.53	&	{81.61}	&	{94.95}	&	90.07	&	95.04	&	{89.94}	&	80.66	&	85.84	&		&	86.31		\\																		
      & MCFA & 78.00
      & 91.00& 	78.00& 	79.00& 	74.00& 	82.00& 	80.00& 	85.00& 	85.00& 	78.00& 	88.00& 	77.00& 	79.00& 	93.00& 	{91.00} & 	94.00& 	89.00& 	{82.00}& 	87.00 &    & 83.63

	\\		
&  {GNAS} & -&	-&	-&	-&	-&	-&	-&	-&	-&	-&	-&	-&	-&	-&	-&	-&	-&	-&	-&	& 86.37

\\		
	& {PS-MCNN-LC} & 84.60&	94.47&	83.51&	82.01&	77.90&	94.97&	87.52&	87.50&	88.81&	84.42&	92.70&	79.65&	83.35&	95.54&	91.21&	95.70&	90.92&	82.18&	86.88&	& 87.67
\\															
	    & DMM-CNN	&	\textbf{84.45}	&	\textbf{94.46}	&	\textbf{83.67}	&	\textbf{82.48}	&	\textbf{76.94}	&	\textbf{91.86}	&	\textbf{84.51}	&	\textbf{86.30}	&	\textbf{86.44}	&	\textbf{82.99}	&	\textbf{92.24}	&	\textbf{79.20}	&	\textbf{79.87}	&	\textbf{94.14}	&	\textbf{90.84}	&	\textbf{95.11}	&	\textbf{89.47}	&	\textbf{81.28}	&	\textbf{88.94}	&		&	\textbf{86.56}	\\																			

		\bottomrule  
	\end{tabular}}
	\label{tab:CelebA}
\end{table*}

\subsection{Comparison with State-of-the-art FAC Methods}
In this subsection, we compare the performance of the proposed DMM-CNN method with several state-of-the-art FAC methods, including (1) PANDA \cite{b11}, which uses part-based models to extract features and SVMs as classifiers; (2) LNets+ANet \cite{b27}, which cascades two localization networks and one attribute network, and uses one SVM classifier for each attribute; (3) MOON \cite{b19}, a novel mixed objective optimization network which addresses the multi-label imbalance problem; (4) NSA (with the median rule) \cite{b14}, which uses segment-based methods for FAC; (5) MCNN-AUX \cite{b20}, which divides 40 attributes into nine groups according to attribute locations;
(6) MCFA \cite{b18}, our previous work which exploits the inherent dependencies between FAC and auxiliary tasks (face detection and FLD). Note that
the accuracy obtained by MOON is not given on the LFWA dataset, since MOON does not report the results on LFWA.
(7) GNAS \cite{Huang}, which proposes an efficient greedy neural architecture search method to automatically learn the multi-attribute deep network architecture.
(8) AW-CNN \cite{b30}, which develops a novel adaptively weighted multi-task
deep convolutional neural network to predict person attributes.
(9) PS-MCNN-LC \cite{Cao}, which introduces a partially shared multi-task network by exploiting both identity information and attribute relationship.

Table \ref{tab:CelebA} shows that DMM-CNN outperforms these competing methods and achieves the mean accuracy of 91.70\% (86.56\%) on CelebA (LFWA). Compared with PANDA and LNets+ANet which use per attribute SVM classifiers, DMM-CNN achieves superior performance by taking advantage of multi-label learning. Our DMM-CNN also achieves better performance than MCNN-AUX, NSA and MOON. It is worth pointing out that our method leverages only two groups of attributes  (i.e., objective and  subjective attributes) while MCNN-AUX employs nine groups of attributes. DMM-CNN is able to achieve higher accuracy than MCNN-AUX, even with fewer attribute groups.
DMM-CNN outperforms MCFA by large margins, which validates the effectiveness of using more facial landmarks information and our attribute grouping mechanism.

The proposed DMM-CNN method achieves similar accuracy with MCNN-AUX on LFWA. DMM-CNN achieves the highest accuracy for 20 attributes among all the 40 attributes, where the performance of subjective attributes (such as ``Pointy Nose", ``Smiling" and ``Bushy Eyebrows") is significantly improved compared with the competing methods.
The proposed DMM-CNN method achieves better performance than GNAS in terms of average recognition rate on both the CelebA and LFWA datasets. This can be ascribed to the effectiveness of the proposed multi-task multi-label learning framework, where two different network architectures are respectively designed to extract features for classifying objective and subjective attributes. Unlike DMM-CNN that manually designs the network architectures, GNAS
automatically discovers the tree-like deep neural network architecture for multi-attribute learning. Therefore, the training process of GNAS is relatively time-consuming.  Compared with AW-CNN, the proposed DMM-CNN method obtains similar accuracy. Different from AW-CNN that predicts multiple person attributes by using the framework of multi-task learning (identifying an attribute is viewed as a single task), the proposed  method jointly learns two closely-related tasks (i.e., FLD and FAC). Note that, the proposed DMM-CNN method achieves
worse performance than PS-MCNN-LC on both the CelebA and LFWA datasets.
PS-MCNN-LC designs a shared network (SNet) to learn the shared features
for different groups of attributes, while adopting the task specific networks (TSNets) for each group of attributes from low-level layers to high-level layers.  However, PS-MCNN-LC takes advantage of the Local Constraint Loss (LCLoss), which requires the face identity as an additional attribute. Moreover, the numbers of channels in SNet and TSNets also need to be carefully chosen to ensure the final performance.
On the whole, the performance comparison between all the competing methods shows the effectiveness of the proposed method.

\section{Conclusion}
In this paper, we propose a novel deep multi-task multi-label CNN method (DMM-CNN) for FAC. DMM-CNN effectively improves the performance of FAC by jointly performing the tasks of FAC and FLD. Based on the division of objective and subjective attributes, different network architectures and a novel dynamic weighting scheme
are adopted for dealing with the diverse learning complexities of facial attributes. For multi-label learning, an adaptive thresholding strategy is developed to alleviate the problem of class imbalance.
Experiments on the public CelebA and LFWA datasets have demonstrated that DMM-CNN achieves superior performance compared with several state-of-the-art FAC methods.

\section*{Acknowledgements}
\small This work was supported by the National Key R\&D Program of China under Grant 2017YFB1302400, by the National Natural Science Foundation of China under Grants 61571379, U1605252, 61872307, by the Natural Science Foundation of Fujian Province of China under Grants 2017J01127 and 2018J01576.

%

%



\ifCLASSOPTIONcaptionsoff
  \newpage
\fi



%

\vspace{-10 mm}
%
\vspace{2 mm}



\vfill


\end{document}